\documentclass[10pt,conference]{IEEEtran}
\IEEEoverridecommandlockouts
\usepackage{cite}
\usepackage{amsmath,amssymb,amsfonts}
\usepackage{algorithmicx}
\usepackage{algpseudocode}
\usepackage{tcolorbox}
\tcbuselibrary{skins}
\usepackage{graphicx}
\usepackage{textcomp}
\usepackage{xcolor}
\usepackage{booktabs}
\usepackage{tabularx}
\usepackage{comment}
\usepackage{subcaption}
\def\BibTeX{{\rm B\kern-.05em{\sc i\kern-.025em b}\kern-.08em
    T\kern-.1667em\lower.7ex\hbox{E}\kern-.125emX}}
\begin{document}

\title{Adaptive Targeted Dynamic Chunking for Tokenization-Free Hierarchical Model
}
\author{\IEEEauthorblockN{1\textsuperscript{st} Thang Dang}
\IEEEauthorblockA{\textit{Fujitsu Research of America} \\
\textit{Pittsburgh, PA, USA}\\
tdang@fujitsu.com}
\and
\IEEEauthorblockN{2\textsuperscript{nd} Akira Nakagawa}
\IEEEauthorblockA{\textit{Fujitsu Limited} \\
\textit{Kawasaki, Kanagawa, Japan}\\
anaka@fujitsu.com}
\and
\IEEEauthorblockN{3\textsuperscript{rd} Kenichi Kobayashi}
\IEEEauthorblockA{\textit{Fujitsu Limited} \\
\textit{Kawasaki, Kanagawa, Japan}\\
kenichi@fujitsu.com}
\and
\IEEEauthorblockN{4\textsuperscript{th} Koichi Shirahata}
\IEEEauthorblockA{\textit{Fujitsu Limited} \\
\textit{Kawasaki, Kanagawa, Japan}\\
k.shirahata@fujitsu.com}
}

\maketitle
\begin{abstract}
Tokenization-free hierarchical models are emerging as a promising alternative to traditional Large Language Models (LLMs), addressing inherent preprocessing issues such as vocabulary design complexity, out-of-vocabulary (OOV) errors, and language-specific constraints. However, a significant challenge in these byte-level methods is the optimization of the compression ratio, a critical factor that dictates model performance for processing bytes data via chunks. In this paper, we propose Adaptive Targeted Dynamic Chunking (ATDC), a novel byte-compression control mechanism designed to enhance the effectiveness of dynamic chunking within hierarchical architectures.
Our approach utilizes curriculum learning to progressively adjust the compression ratio during training, transitioning from low to high compression to stabilize the learning process. We provide an analysis establishing the relationship between the target compression ratio and Bytes-Per-Innermost-Chunk (BPIC), allowing for tracking of chunk-size evolution throughout the training phase.
Evaluations conducted on the FineWeb-Edu 100B dataset demonstrate that hierarchical models equipped with ATDC achieve competitive Bits-Per-Byte (BPB) performance compared to conventional baselines operating at both byte and token levels. Furthermore, the proposed method exhibits more stable training dynamics and superior final performance across diverse downstream tasks compared to models using fixed compression ratios, while maintaining the inherent robustness and flexibility of byte-level processing.

\end{abstract}


\section{Introduction}

The development of Large Language Models (LLMs)~\cite{dang2022regularizing, kasagi2021efficient} has traditionally depended on subword tokenization techniques, including Byte-Pair Encoding (BPE)~\cite{kudo2018sentencepiece, schmidt2025boundless}, WordPiece~\cite{song2021fast}, and hybrid Byte-Word approaches~\cite{neitemeier2025hierarchical} applied during preprocessing. Although these methods have proven effective, they impose notable limitations. 
Tokenization demands intricate vocabulary construction, which frequently results in out-of-vocabulary (OOV) problems and reduced resilience to typographical errors or morphological diversity~\cite{ahia2023do}. Moreover, tokenizers exhibit pronounced bias toward high-resource languages~\cite{petrov2023language}, exacerbating performance disparities in multilingual contexts. These intrinsic drawbacks, often termed the ``overheads" of tokenization, have driven interest in tokenization-free architectures~\cite{pagnoni2025byte} that process raw bytes directly.
While byte-level models hold considerable potential, they encounter a core efficiency hurdle: individual byte or meaningful chunks span numerous bytes, necessitating substantial compression to handle extended contexts and control computational overhead effectively. Hierarchical designs, such as H-Net~\cite{nawrot2022hierarchical}, mitigate this through mechanisms like dynamic chunking~\cite{hwang2025dynamic} or patching~\cite{pagnoni2025byte} to form higher-level abstractions from byte sequences. Nevertheless, determining and regulating the compression ratio in these systems remains predominantly heuristic and suboptimal. Existing approaches typically employ static compression rates (for example, compression rate N is equal to 6 indicates that average of 6 bytes are expected to be compressed into a single chunk) that fail to adapt to evolving data complexity throughout training, often causing unstable optimization or suboptimal dynamic compression.
In this work, we address these shortcomings by proposing Adaptive Targeted Dynamic Chunking (ATDC). Our approach is motivated by the insight that a model's compression capability should adapt in tandem with its growing comprehension of the data distribution. We conceptualize compression management as a form of curriculum learning, wherein the method gradually modulates the compression ratio and targeted chunk sizes. This enables a smooth progression from low-compression (fine-grained) processing to high-compression (coarse-grained) regimes as training advances.

The primary contributions of this work are as follows:
\begin{itemize}
\item A Novel Compression Control Method: We propose ATDC, which dynamically regulates byte compression in hierarchical  architectures.

\item Indicator Framework: We provide a formal analysis establishing the relationship between a target compression ratio and the Bytes-Per-Innermost-Chunk (BPIC).

\item Empirical Validation: Through extensive training on 100B tokens of the FineWeb-Edu~\cite{penedo2024fineweb} dataset, we demonstrate that byte-level models using ATDC achieve bits-per-byte performance competitive with both byte-level H-Net~\cite{hwang2025dynamic} and token-level Llama3.2~\cite{grattafiori2024llama}-based models.

\item Textual Robustness: We demonstrate that our adaptive approach leads to increased robustness against HellaSwag textual perturbations across a variety of tasks compared to models with a fixed compression rate.
\end{itemize}

\section{Related works}
Tokenization has become a significant limitation in modern Large Language Models, particularly due to its sensitivity to spelling variations, typos, capitalization, and morphological changes~\cite{xue2022byt5tokenfreefuturepretrained}. Byte-level modeling, which directly processes raw bytes, offers a promising alternative by eliminating many of these preprocessing assumptions.

MegaByte~\cite{yu2023megabyte} introduced a multiscale decoder capable of modeling extremely long byte sequences (over 1 million bytes). However, its reliance on fixed-size patches and simple byte embedding concatenation limits flexibility.

MambaByte~\cite{wang2024mambabyte} advanced this direction by training a Mamba-based~\cite{gu2022efficiently} model directly on raw bytes, leveraging its efficient fixed-size state. Still, compressing all contextual information into a single fixed hidden state remains a potential bottleneck.

More recently, the Byte Latent Transformer (BLT)~\cite{pagnoni2025byte} proposed a dynamic patching strategy that groups bytes according to next-byte prediction entropy. This approach adaptively assigns more computation to complex regions while using less for predictable sequences. However, BLT’s entropy-based method can suffer from instability or “drift,” especially on highly repetitive text such as in multiple-choice reasoning tasks.

Dynamic Chunking~\cite{hwang2025dynamic} utilizes an end-to-end routing mechanism and exponential moving averages to learn chunk boundaries differentiably. However, its reliance on a fixed compression ratio introduces sensitive, manually-tuned hyperparameters. While H-Net++~\cite{zakershahrak2025h} attempts to improve this via a lightweight Transformer context-mixer, its evaluation on limited data (1.4B tokens) and smaller model scales prevents a comprehensive performance comparison.


Despite these advancements, Dynamic Chunking is limited by its use of a fixed compression rate for chunk size control. Such a mechanism results in highly empirical hyperparameters that often demand manual tuning. Our work is motivated by the need to move beyond these manual constraints, a goal we achieve through our proposed ATDC method.

\section{Hierarchical Models}

H-Net~\cite{hwang2025dynamic} is a tokenizer-free hierarchical language model that directly processes raw byte sequences ($\mathcal{V}vocab = 256$). The primary innovation is Dynamic Chunking (DC), a mechanism that identifies semantic boundaries in byte sequences to facilitate efficient hierarchical processing. The general workflow is: $\text{Input Bytes} \to \text{Encoder} \to \text{Routing Module} \to \text{Chunking} \to \text{Main Network} \to \text{Dechunking} \to \text{Decoder} \to \text{Output}$.

\subsection{Hierarchical Decomposition}
Given an input sequence $\mathbf{x} = (x_1, x_2, \dots, x_L)$ of length $L$, H-Net constructs a hierarchical representation through recursive compression. At any stage $s$, the process is defined by:
\begin{itemize}
    \item \textbf{Boundary Prediction:} A routing module predicts discrete decisions $\mathbf{b}^{(s)} \in \{0,1\}^{L_s}$ and continuous probabilities $\mathbf{p}^{(s)} \in [0,1]^{L_s}$ from hidden states $\mathbf{h}^{(s)} \in \mathbb{R}^{L_s \times d_s}$.
    \item \textbf{Compression:} The sequence length for the subsequent stage is determined by the number of boundaries, $M_{s+1} = \|\mathbf{b}^{(s)}\|_1$.
\end{itemize}
For a $K$-stage hierarchy with downsampling factors $\{n_1, n_2, \dots, n_K\}$, the total compression ratio is given by $\prod_{i=1}^K n_i$. For example, a two-stage model with $N_1 = N_2 = 3$ achieves approximately $9\times$ compression.

\subsection{Boundary Detection via Semantic Discontinuity}
The routing module identifies boundaries by quantifying the semantic shift between consecutive hidden states using a learned cosine similarity:
\begin{equation}
    \sigma_t = \frac{\langle Q(\mathbf{h}_{t-1}), K(\mathbf{h}_t) \rangle}{\| Q(\mathbf{h}_{t-1}) \| \cdot \| K(\mathbf{h}_t) \|}
\end{equation}
where $Q$ and $K$ are learnable linear projections initialized as identity matrices. The boundary probability $p_t$ and discrete decision $b_t$ are then derived as:
\begin{equation}
    p_t = \text{clip}\left(\frac{1 - \sigma_t}{2}, 0, 1\right), \quad b_t = \text{argmax}(1 - p_t, p_t)
\end{equation}
High similarity ($\sigma_t \approx 1$) signals semantic continuity ($p_t \approx 0$), while low similarity ($\sigma_t \approx -1$) triggers a boundary ($p_t \approx 1$). This formulation learns a data-dependent metric space where cosine distance aligns with natural semantic transitions.

\subsection{Component-wise Analysis}

\subsubsection{Encoder and Routing}
The encoder $\mathcal{E}_{\text{enc}}$ contextualizes byte embeddings using $L$ layers alternating between Mamba-2 State-Space Models (SSMs)~\cite{pmlr-v235-dao24a}, multi-head attention, and SwiGLU~\cite{shazeer2020glu} feed-forward networks. The routing function $R$ maps these states to probabilities $\mathbf{p} = R(\mathbf{h}_{\text{enc}}; \mathbf{h}_{\text{route}})$. To allow gradient flow through the discrete argmax, we use a Straight-Through Estimator (STE)~\cite{bengio2013estimating}, treating the discretization as an identity during backpropagation:
\begin{equation}
    \frac{\partial L}{\partial \mathbf{h}} = \frac{\partial L}{\partial \mathbf{p}} \cdot \frac{\partial \mathbf{p}}{\partial \mathbf{h}}
\end{equation}

\subsubsection{Dechunking via EMA}
The dechunking operator $\mathcal{D}$ reconstructs the original sequence length using an Exponential Moving Average (EMA)~\cite{brotons2024exponential} recurrence:
\begin{equation}
    \mathbf{y}_t = p_t \cdot \mathbf{c}_t + (1 - p_t) \cdot \mathbf{y}_{t-1}
\end{equation}
where $\mathbf{c}_t$ represents the chunk value at boundaries. This operation is causal, differentiable, and can be efficiently computed via parallel scan kernels by reformulating the recurrence as a linear SSM. 

\subsubsection{Gated Residual Connections}
To ensure stable hierarchical training, we utilize gated residual connections:
\begin{equation}
    \mathbf{h}_{\text{out}} = \mathbf{h}_{\text{dechunk}} \cdot \text{STE}(\mathbf{p}) + W_{\text{res}}(\mathbf{h}_{\text{enc}})
\end{equation}
where $W_{\text{res}}$ is a zero-initialized projection. This creates an identity-path initialization, allowing the model to gradually learn the hierarchical features without destabilizing initial training.

\subsection{Main Network: Recursive Hierarchical Processing}
The main network $\mathcal{M}_{\text{main}}$ processes the compressed latent representations $\mathbf{h}_{\text{chunk}}$ generated by the Dynamic Chunking (DC) mechanism. Its architecture can be configured in two primary modes depending on the hierarchical depth $K$:

\begin{enumerate}
    \item \textbf{Recursive H-Net Stage:} For $K > 1$, the main network recursively applies another H-Net stage. This creates a multi-scale hierarchy where each subsequent layer operates on a progressively coarser granularity, effectively building a tree-like computation graph.
    \item \textbf{Isotropic Model:} In the final stage, the compressed sequence is processed by a standard sequence model (e.g., a deep Transformer or a Mamba-2 backbone). Since the sequence length is reduced from $L$ to $M \approx L/r$, the attention mechanism's quadratic complexity $\mathcal{O}(M^2)$ or the SSM's linear complexity $\mathcal{O}(M)$ is significantly mitigated.
\end{enumerate}

\paragraph{Mathematical Formulation}
For a $K$-stage hierarchy, let $\mathcal{M}_{\text{main}}^{(k)}$ denote the processing block at stage $k$. The state transition is defined as:
\begin{equation}
    \mathbf{h}^{(k)} = 
    \begin{cases} 
    \mathcal{M}_{\text{base}}(\mathbf{h}^{(k-1)}) & k = K \\
    \mathcal{M}_{\text{main}}^{(k)}(\mathbf{h}^{(k-1)}) & k < K
    \end{cases}
\end{equation}
This recursive structure allows the model to capture long-range dependencies at the coarsest levels while maintaining the ability to reconstruct fine-grained local details during the dechunking and decoding phases. By bottlenecking the information flow through a learned semantic boundary, the Main Network focuses its parameters on the most salient features of the byte stream.

\section{Adaptive Targeted Dynamic Chunking}

We propose the \textit{Adaptive Targeted Dynamic Chunking} (ATDC) mechanism to dynamically regulate dynamic chunking during training. We retain the original routing formulation from~\cite{hwang2025dynamic} but modify the compression control mechanism. ATDC stabilizes the learning process by employing a scheduled linear growth of the compression factor $N$, supplemented by a performance-driven adaptation trigger and a localized balancing loss to maintain structural integrity.

\subsection{Adaptive Routing Schedule}

The compression factor $N(t)$ dictates the target reduction in sequence length at training step $t$. To facilitate a stable transition from a low-compression (simplified) architecture to a high-capacity state, we implement a multi-phase schedule. During the initial warm-up period ($t < T_{w}$), $N(t)$ is held constant at $N_{init}$ to allow the router to converge on basic patterns. Subsequently, $N(t)$ follows a linear interpolation toward a target value $N_{fnl}$ based on the training progress:

\begin{equation}
\label{alg:N_sched}
N_{sched}(t) = 
\begin{cases} 
N_{init} & t < T_{w} \\
N_{init} + \frac{t - T_{w}}{T - T_{w}} (N_{fnl} - N_{init}) & t \geq T_{w}
\end{cases}
\end{equation}

\begin{tcolorbox}[
  colback=white,
  colframe=black!70!black,
  title={Adaptive Targeted Dynamic Chunking},
  fonttitle=\bfseries,
  sharp corners
]
\begin{algorithmic}[1]
\label{alg:atdc}
\State \textbf{Input:} Steps $T$, Warmup $T_w$, Targets $\mathbf{N}_{init}, \mathbf{N}_{fnl} \in \mathbb{R}^S$, Rate $\gamma$, Threshold $\tau$, Window $W$
\State $t \gets 0, \mathcal{H} \gets \emptyset$ \Comment{Initialize for metric-based adaptation}

\While{$t < T$}
    \Comment{Adaptive Scheduling}
    \If{$t < T_w$}
        \State $\mathbf{N}_{sched} \gets \mathbf{N}_{init}$
    \Else
        \State $\rho \gets \min\left(\frac{t - T_w}{T - T_w}, 1\right)$
        \State $\mathbf{N}_{sched} \gets \mathbf{N}_{init} + \rho \cdot (\mathbf{N}_{fnl} - \mathbf{N}_{init})$
    \EndIf

    \If{$|\mathcal{H}| \geq W$} \Comment{Wait for window to fill}
        \State $\bar{\ell} \gets \text{mean}(\mathcal{H}[-W:])$  \Comment{Average loss}
        \State $\mathbf{N}_{curr} \gets (\bar{\ell} < \tau) ? (\mathbf{N}_{sched} \cdot \gamma) : \mathbf{N}_{sched}$
    \Else
        \State $\mathbf{N}_{curr} \gets \mathbf{N}_{sched}$
    \EndIf

    
    \State $(\mathbf{y}, \{\mathcal{R}_s\}_{s=1}^S) \gets \mathcal{M}(\mathbf{x})$ \Comment{Forward Pass \& Balancing}
    \State $\mathcal{L}_{bal} \gets 0$
    \For{$s \gets 1$ \textbf{to} $S$}
        \State $N_s \gets \mathbf{N}_{curr}[s]$
        \State $\mathcal{L}_{bal} \gets \mathcal{L}_{bal} + \Call{BalancingLoss}{\mathcal{R}_s, N_s}$
    \EndFor
    \State $\mathcal{L}_{total} \gets \text{CrossEntropy}(\mathbf{y}, \mathbf{x}) + \alpha \mathcal{L}_{bal}$

    \State Update $\mathcal{M}$ via $\nabla_{\theta} \mathcal{L}_{total}$; $\mathcal{H}.\text{append}(\mathcal{L}_{total})$
    \State $t \gets t + 1$
\EndWhile

\Statex \hrulefill

\Function{BalancingLoss}{$\text{out}_s$, $N$}
    \State $Y \gets \text{mean}(\text{out}_s\text{.mask})$
    \State $\quad Y' \gets \text{mean}(\text{out}_s\text{.prob})$ 
    \State $\mathcal{L}_{bal} \gets \frac{N}{N - 1} \times \Bigl[ (N - 1)YY' + (1 - Y)(1 - Y') \Bigr]$
    \State \Return $\mathcal{L}_{bal}$
\EndFunction
\end{algorithmic}

\end{tcolorbox}

To further optimize the training trajectory, we introduce a metric-based adaptation. We maintain a history of the language modeling loss within a sliding window $W$. If the windowed average loss $\bar{\ell}$ falls below a predefined threshold $\tau$, indicating the model has reached a sufficient level of proficiency, we apply an adaptation rate $\gamma$ to accelerate capacity acquisition:

\begin{equation}
N(t) = 
\begin{cases} 
N_{sched}(t) \cdot \gamma & \text{if } \bar{\ell} < \tau \\
N_{sched}(t) & \text{otherwise}
\end{cases}
\end{equation}

\subsection{Balancing Loss}

To regularize the routing mechanism and prevent degenerate solutions, and we utilize a balancing loss~\cite{hwang2025dynamic} $\mathcal{L}_{bal}$. This term aligns the router's soft probabilistic outputs $Y'$ with the discrete routing decisions $Y$ actually executed by the model. Given the current compression control $N$, the loss for a sequence of length $L$ is formulated as:

\begin{equation}
\label{eq:load-balance}
\mathcal{L}_{bal} = \left( \frac{N}{N-1} \right) \Big[ (N - 1)YY' + (1 - Y)(1 - Y') \Big]
\end{equation}

where $Y$ represents the fraction of vectors actually selected by the boundary mask, and $Y'$ denotes the average boundary probability:

\begin{equation}
\label{eq:y_fraction}
Y = \frac{1}{L} \sum_{t=1}^{L} b_t, \quad Y' = \frac{1}{L} \sum_{t=1}^{L} p_t
\end{equation}

In this framework, $b_t \in \{0, 1\}$ is the binary boundary mask and $p_t \in [0, 1]$ is the router's predicted probability for the $t$-th vector.

\begin{itemize}
    \item \textbf{The Boundary Penalty $(N-1)YY'$:} This term penalizes the density of boundary assignments. As the target $N$ increases, the model is increasingly incentivized to minimize boundary selections, thereby enforcing higher compression.
    \item \textbf{The Consistency Term $(1 - Y)(1 - Y')$:} This ensures that for vectors not selected as boundaries, the router remains confident in its non-selection (predicting low $p_t$), which stabilizes the gradients for ``skipped'' tokens.
    \item \textbf{Normalization $\frac{N}{N-1}$:} This coefficient ensures that the balancing loss remains numerically stable and comparable in magnitude across different phases of the $N$ schedule, preventing the regularization from dominating the global loss as $N$ grows.
\end{itemize}

\subsection{Compression Control by ATDC and Indicator as BPIC}
Our method ATDC regulates the chunk size of router, via the ratio loss, it acts as a soft constraint on compression, allowing the model to deviate when semantically necessary while maintaining overall efficiency.

\subsubsection{Compression Control and Load Balancing}
Recall $N$ is the target compression ratio and $\rho = 1/N$ be the target boundary fraction. The ATDC mechanism optimizes a ratio loss $\mathcal{L}_{bal}$ defined in (\ref{eq:load-balance}).
The total objective loss function is:
\begin{equation}
\label{eq:loss}
    \mathcal{L}_{\text{total}} = \alpha \mathcal{L}_{bal} + \mathcal{L}_{\text{entropy}}
\end{equation}
where $\alpha$ is a scaling hyperparameter, $\mathcal{L}_{\text{entropy}}$ is a cross-entropy loss of the next bytes prediction in the sequence. This objective ensures that the model maintains the target throughput while using the entropy term to prevent over-confident or uniform boundary predictions.

\subsubsection{Boundary Logic and Segmentation}
The router determines boundaries by measuring the semantic shift between consecutive states $\mathbf{h}_t$ and $\mathbf{h}_{t+1}$ in a learned metric space:
\begin{equation}
\label{eq:boundary_probability}
    p_t = \frac{1}{2} \left( 1 - \frac{Q(\mathbf{h}_t) \cdot K(\mathbf{h}_{t+1})}{\|Q(\mathbf{h}_t)\| \|K(\mathbf{h}_{t+1})\|} \right)
\end{equation}
A discrete boundary mask $b \in \{0, 1\}$ is generated by $b_t = \mathbb{I}(p_t > 0.5)$. Sequential positions between active indices in $b$ are grouped into chunks. We monitor the empirical compression performance by calculating \textbf{Bytes Per Innermost Chunk (BPIC)}:
\begin{equation}
    \text{BPIC} = \frac{1}{M} \sum_{i=1}^M \text{size}(C_i)
\end{equation}
where $M$ is the total number of chunks and $\text{size}(C_i)$ is the length of the $i$-th segment.

\section{Implementation}
\label{experiments}
\subsection{Model}
We compare against the byte-level H-Net model~\cite{hwang2025dynamic}. We additionally compare against token-level architecture Llama3.2~\cite{grattafiori2024llama} model. H-Net supports K-stage, for computational cost reason, we only use one stage (K=1) for all H-Net models. The isotropic model or main network of H-Net is Transformer architecture (vanilla Llama architecture).
\subsection{Cross-Model Evaluation Metrics}
We use 100B tokens subset sampled from the English Fineweb-edu~\cite{penedo2024fineweb} dataset. To ensure a fair comparison between the byte-level model and the token-level baseline, we normalize all performance results to Bits-Per-Byte (BPB). While the token-level model operates on a sequence length of $L_{token} = 1732$, the byte-level model processes $L_{byte} = 8192$ UTF-8 encoded bytes, representing an equivalent volume of raw information. 


The token-level perplexity ($PPL$) is converted to BPB using the equation: $BPB = (\log_2(PPL) \cdot L_{token})/ L_{byte}$

This normalization accounts for the tokenizer's compression ratio (approximately $4.73$ bytes per token) and allows for a vocabulary-agnostic comparison of the models' generative efficiency.

\subsection{Hyperparameters}

The performance of the ATDC mechanism relies on the selection of the adaptation threshold $\tau$ and the acceleration rate $\gamma$. These parameters govern the trade-off between predictive accuracy and computational efficiency.

\subsubsection{Adaptive targeted N}
We utilize the same settings as ~\cite{hwang2025dynamic} for all H-Net baseline models, with $N=6.0$. For the ADTC, we set $N_{init}=5.0$ and $N_{fnl}=6.5$. To maintain training stability, we set the warm-up period to $T_{w}=60\%T$. Following this, adaptive scheduling is deployed to automatically scale up $N_{sched}$ using metric-based adaptation (\ref{alg:N_sched}).
\subsubsection{Selection of Adaptation Threshold ($\tau$)}

The threshold $\tau$ serves as a ``proficiency trigger.'' It defines the maximum allowable language modeling loss at which the model is deemed stable enough to handle increased compression. 

\begin{itemize}
    \item \textbf{Empirical Estimation:} To determine $\tau$, we first conduct a baseline training run with a static $N_{init}$. $\tau$ is set to the loss value achieved at the end of the first epoch or the point where the loss curve begins to plateau. Our empirical $\tau$ value is 0.05.
    \item \textbf{Sensitivity:} Setting $\tau$ too high may result in premature compression and routing instability. Conversely, a $\tau$ that is too low may never trigger the adaptation.
\end{itemize}

\subsubsection{Selection of Adaptation Rate ($\gamma$)}

The rate $\gamma$ determines the magnitude of the opportunistic boost in the compression factor. In our experiments, we fixed $\gamma$ is equal $1.05$ (5$\%$ boost in compression). Because $\gamma$ modifies $N(t)$, which scales $\mathcal{L}_{bal}$, we use a windowed average $\bar{\ell}$ over $W$ (100) steps to ensure the trigger loss changing is based on sustained trends rather than batch-level.

\subsubsection{Scaling hyperparameter of loss function ($\alpha$)}

Purpose: Balances the two objectives $\mathcal{L}_{bal}$ and $\mathcal{L}_{\text{entropy}}$ in (\ref{eq:loss}):
\begin{itemize}
\item $\alpha$ too small: Model ignores load balancing → poor chunking, inefficient hierarchical processing.
\item $\alpha$ too large: Model focuses too much on balanced chunks → worse language modeling performance.
\item $\alpha$ = 0.03: Empirically optimal value that maintains good language modeling while encouraging efficient dynamic chunking.
\end{itemize}

\subsubsection{Others}
We utilized learning rate modulation~\cite{hwang2025dynamic} scaled according to model size. Specifically, the peak learning rates were set as follows: $7.5 \times 10^{-4}$ for the 350M H-Net model, $6.25 \times 10^{-4}$ for the 680M H-Net and 770M Llama 3.2 models, and $5.0 \times 10^{-4}$ for the 1.3B H-Net and 1.5B Llama 3.2 models. Training was conducted using the AdamW optimizer with Decoupled Weight Decay Regularization~\cite{loshchilov2017decoupled} and a Warmup-Stable-Decay (WSD)~\cite{hu2024minicpm} scheduler. This schedule included a $10\%$ initial warmup phase and a $20\%$ inverse-square-root decay phase~\cite{ibrahim2024simple}. We employed a batch size of 256 across 8 NVIDIA H200 GPUs, resulting in approximately 2M bytes per gradient update.

\subsection{Baselines and Metrics}

To evaluate the ATDC method, we compare it against the following configurations:
\begin{itemize}
\item \textbf{Static Routing:} The compression ratio $N$ remains fixed at $N_{init}$ throughout the entire training process.
\item \textbf{Adaptive Scheduling:} The value of $N$ follows the ATDC, metric-based adjustments used in our approach.
\item \textbf{Token-level Reference:} We train and evaluate a token-level Llama 3.2 model to provide a performance benchmark against standard tokenization methods. Because token-level models do not utilize the same byte-level compression metrics, we omit the Llama 3.2 validation BPB from our figures to maintain consistency across the comparative analysis of BPIC and N.
\end{itemize}

\subsection{Results}

\begin{figure}
    \centering
    \includegraphics[width=0.9\linewidth]{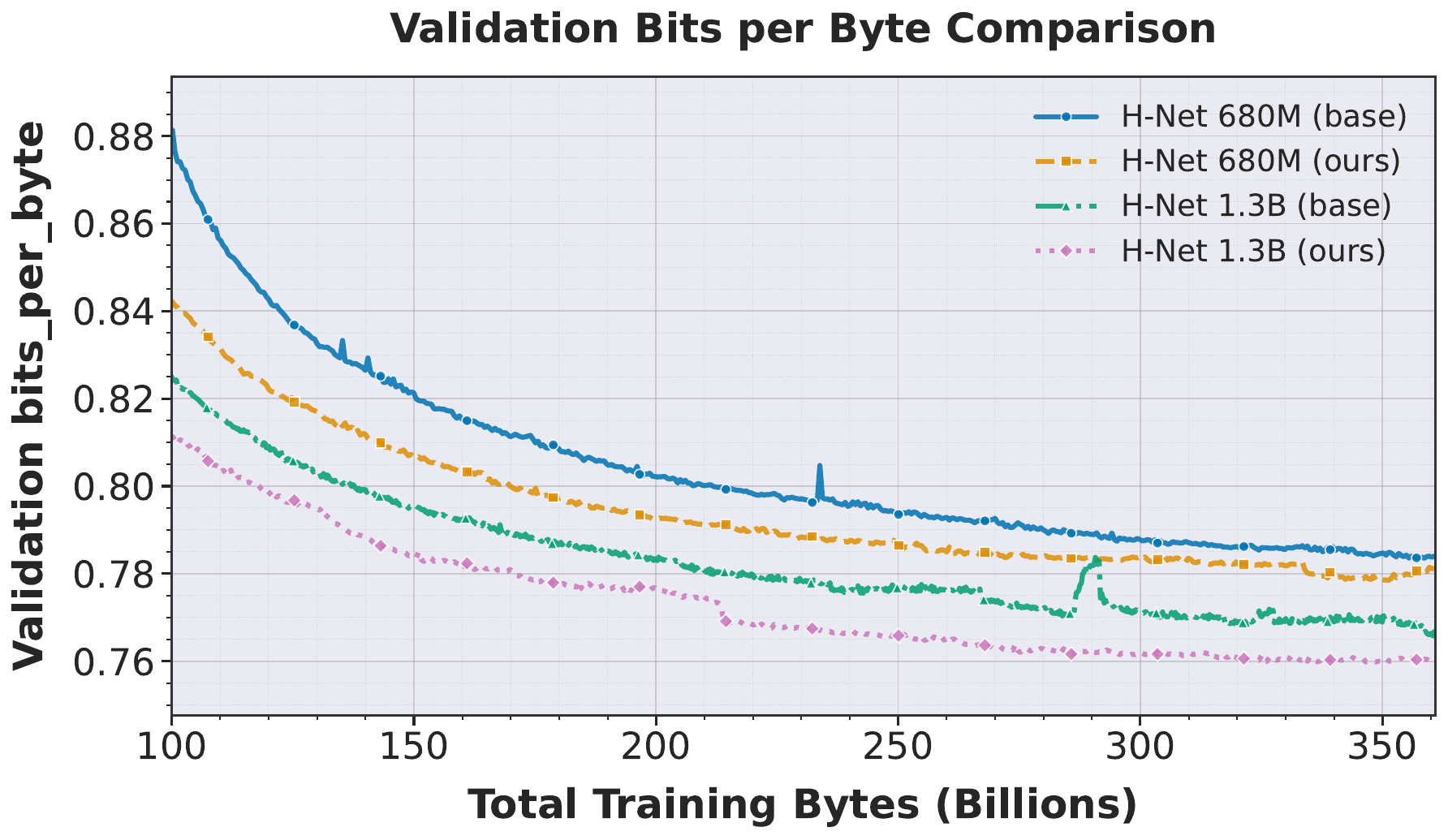}
    \caption{Validation BPB.}
    \label{fig:val_bpb}
\end{figure}


\begin{figure*}[t]
    \centering
    
    \begin{subfigure}[b]{0.495\textwidth}
        \centering
        \includegraphics[width=\linewidth]{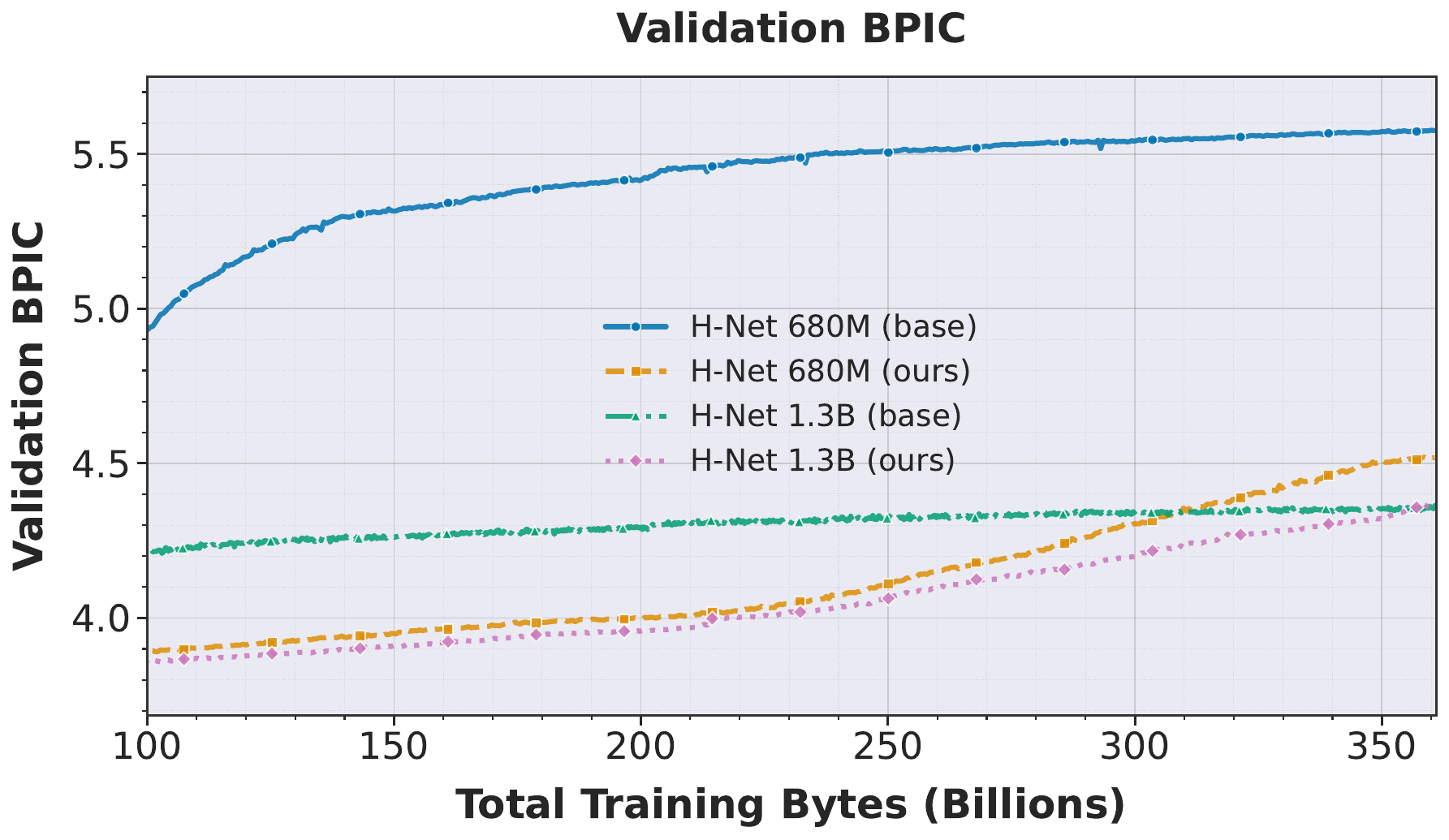}
        \caption{Comparison on BPIC.}
        \label{fig:val_bpic}
    \end{subfigure}%
    \hfill   
    \begin{subfigure}[b]{0.495\textwidth}
        \centering
        \includegraphics[width=\linewidth]{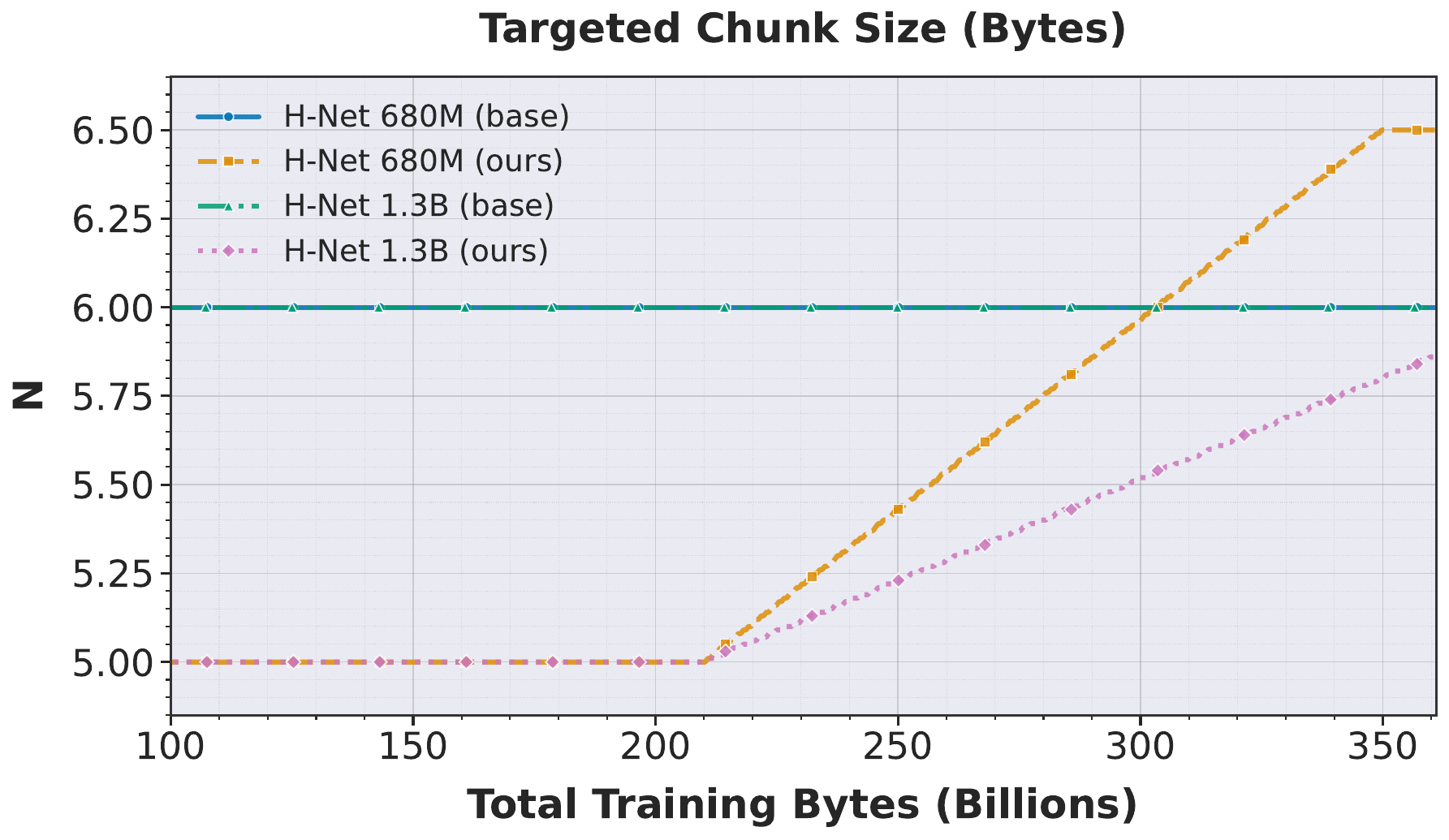}
        \caption{Rate compression control $N$.}
        \label{fig:val_adaptive}
    \end{subfigure}
    
    \caption{Validation BPIC curves to track the N values changing during training phase.}
    \label{fig:bpic_N}
\end{figure*}

Table \ref{tab:bpb_model_comparison} presents a comprehensive evaluation of our proposed method compared to the token-level Llama3.2 and a byte-level H-Net baseline models. The performance is measured using validation BPB (as Fig.~\ref{fig:val_bpb}) for language modeling quality and zero-shot accuracy across seven standard benchmarks.
A primary observation is that byte-level models consistently outperform their token-based counterparts (Llama3.2) across all parameter scales, despite having slightly fewer parameters in some configurations. Notably, the H-Net 680M (ours) achieves an average accuracy of 50.7\%, significantly surpassing the Llama3.2 770M (42.1\%) and even the larger Llama3.2 1.5B (48.5\%). This suggests that direct byte-level modeling captures linguistic nuances and structural information that tokenization-based preprocessing may obscure.
Our implementation consistently achieves the lowest (best) BPB scores in every category, reaching a minimum of 0.760 at the 1.3B parameter scale. 

In terms of zero-shot reasoning: the ADTC outperforms the base H-Net model in nearly every metric, particularly in high-stakes reasoning tasks like ARC-Challenge (ARC-C) and PIQA.



\begin{table*}[t]
\centering
\caption{Validation Bits-per-byte (BPB) and Zero-shot performance comparison across multiple benchmarks}
\label{tab:bpb_model_comparison}
\small 
\begin{tabular*}{\textwidth}{@{\extracolsep{\fill}}lccccccccccc}
\toprule
\textbf{MODEL} & \textbf{INPUT} & \textbf{BPB↓} & \textbf{BPIC} & \textbf{ARC-C↑} & \textbf{ARC-E↑} & \textbf{HELLA↑} & \textbf{LAMB↑} & \textbf{OPEN↑} & \textbf{PIQA↑} & \textbf{WINO↑} & \textbf{AVG↑}  \\
\midrule

Llama3.2 770M & Token & 0.889 & NA & 28.1 & 53.7 & 38.3 & 26.7 & 32.4 & 64.5 & 51.0 & 42.1 \\
H-Net 680M (base) & Byte & 0.783 & 5.58 & 33.4 & 60.1 & 52.3 & 40.9 & \textbf{39.2} & 70.9 & 55.7 & 50.4 \\
H-Net 680M (ours) & Byte & \textbf{0.778} & 4.52 & \textbf{33.9} & \textbf{61.2} & \textbf{52.4} & \textbf{42.1} & 37.2 & \textbf{71.7} & \textbf{56.1} & \textbf{50.7} \\
\hline \hline
Llama3.2 1.5B & Token & 0.800 & NA & 33.1 & 60.3 & 49.0 & 37.3 & 34.8 & 70.2 & 54.7 & 48.5 \\
    H-Net 1.3B (base) & Byte & 0.766 & 4.35 & 34.4 & 58.7 & \textbf{55.0} & 43.4 & \textbf{38.4} & 70.7 & \textbf{56.8} & 51.1 \\
H-Net 1.3B (ours) & Byte & \textbf{0.760} &4.37 & \textbf{36.9} & \textbf{60.5} & 54.9 & \textbf{44.4} & 37.4 & \textbf{71.5} & 56.2 & \textbf{51.7} \\
\bottomrule
\end{tabular*}
\end{table*}

\textbf{Robustness to Textual Perturbations.} To evaluate the structural resilience of our proposed method, we conducted a stress test on the HellaSwag benchmark using five types of textual perturbations: Antspeak (added spacing), Drop (character deletion), RandomCase, Repeat (character duplication), and Uppercase. The results, summarized in Table \ref{tab:hellag_perturbation}, highlight a significant gap in robustness between token-based and byte-level architectures.

\textbf{Resilience Against Out-of-Distribution Noise.}
The token-based Llama 3.2 models exhibit a sharp decline in performance when faced with even minor character-level noise. For instance, at the 1.5B scale, Llama 3.2’s accuracy drops to an average of 27.7\% under perturbation. In contrast, our H-Net 1.3B (ours) maintains an average accuracy of 31.6\%. This confirms that subword tokenizers are highly sensitive to spelling variations and formatting changes, as these perturbations often result in "unknown" tokens or fragmented subword sequences that disrupt the model's semantic understanding.

\textbf{Comparison of Byte-Level Strategies.}
While both byte-level models (base and ours) significantly outperform the token-based baseline, our method consistently achieves the highest accuracy across nearly all noise categories.

\textbf{Case Sensitivity}. Our model shows a particular advantage in the UPPERCASE and RandomCase categories. Specifically, at the 1.3B scale, our method achieves 39.7\% on uppercase text, outperforming the base byte model and vastly exceeding Llama 3.2 (29.6\%).

\textbf{Structural Consistency.} In the Antspeak and Repeat, which simulate common typos and stylized web text, our method demonstrates superior stability. This suggests that the adaptive thresholding in our dynamic chunking mechanism is better at identifying core morphological units. 

\begin{table*}[t]
    \centering
    \caption{The Evaluation on HellaSwag with textual perturbations}
    \label{tab:hellag_perturbation}
    \begin{tabular*}{\textwidth}{@{\extracolsep{\fill}}lccccccc}
        \toprule
        \textbf{MODEL} & \textbf{INPUT} & \textbf{ANTSPEAK} & \textbf{DROP} & \textbf{RANDOMCASE} & \textbf{REPEAT} & \textbf{UPPERCASE} & \textbf{AVERAGE-ACC↑} \\
        \midrule

        Llama3.2 770M     &Token    & 28.0 & 26.0 & 25.5 & 26.0 & 27.1 & 26.5 \\
        H-Net 680M (base)&Byte     & 30.0 & 27.9 & 27.5 & \textbf{29.2} & 37.7 & 30.5 \\
        H-Net 680M (ours)&Byte     & \textbf{30.6} & \textbf{28.0} & \textbf{27.8} & 29.0 & \textbf{38.0} & \textbf{30.7} \\
        \hline \hline
        Llama3.2 1.5B     &Token    & 29.7 & 26.6 & 25.7 & 26.9 & 29.6 & 27.7 \\
        H-Net 1.3B (base)&Byte     & 31.2 & 28.2 & 26.9 & 29.0 & 38.8 & 30.8 \\
        H-Net 1.3B (ours)&Byte     & \textbf{31.3} & \textbf{28.7} & \textbf{28.3} & \textbf{29.9} & \textbf{39.7} & \textbf{31.6} \\
        \bottomrule
    \end{tabular*}
\end{table*}

\textbf{Visualization of Chunking Boundaries.}

\begin{figure*}
    \centering
    \includegraphics[width=0.9\linewidth]{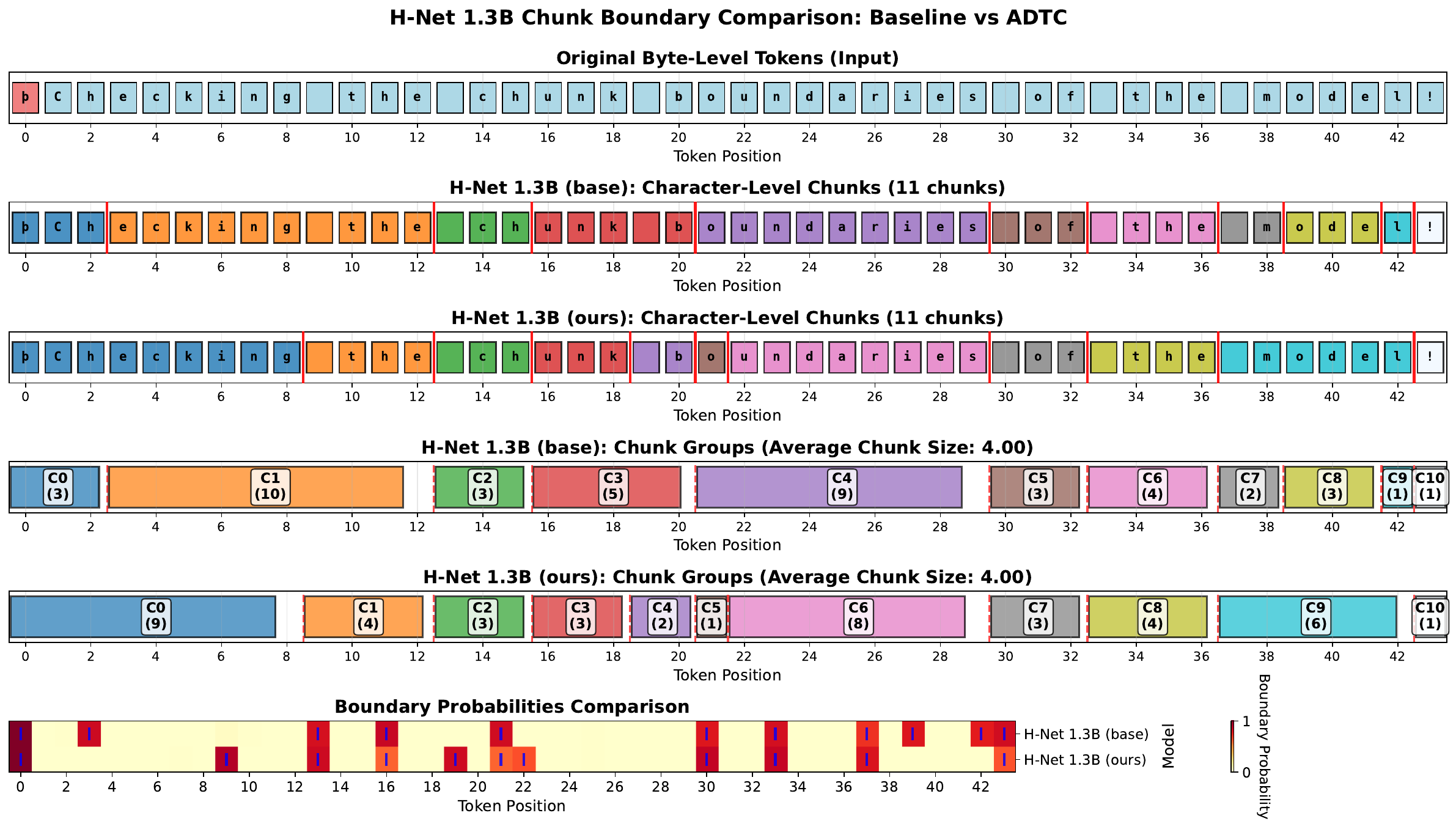}
    \caption{Chunk Boundary Visualization of H-Net 1.3B.}
    \label{fig:visual_chunking}
\end{figure*}

We utilize our largest model variant (1.3B parameters) to analyze chunk boundaries and compare our proposed method against the conventional fixed-rate approach. As illustrated in Fig. \ref{fig:visual_chunking}, the H-Net model determines these boundaries by calculating the boundary probability via (\ref{eq:boundary_probability}), which subsequently dictates the masking and chunk compression strategy. The visualization reveals that the conventional method (second row) produces less semantically meaningful segmentations compared to our method (third row). For instance, the conventional approach splits the word ``checking" across two separate chunks, $C_0$ (``ch") and $C_1$ (``ecking"), while simultaneously grouping the word ``the" into $C_1$. In contrast, our method maintains the integrity of the word ``checking" within a single chunk, $C_0$. A similar pattern is observed with the word ``model": the conventional method fragments it into three distinct chunks ($C_7, C_8, C_9$), whereas our approach successfully preserves it as a single unit in $C_9$. These observations suggest that our adaptive mechanism better captures linguistic structures than fixed-rate alternatives.

\textbf{Indicator BPIC and adaptive N analysis.}

Fig.~\ref{fig:bpic_N} illustrates the validation BPIC curves and the corresponding shifts in the adaptive $N$ value after 200B bytes. These indicators allow us to verify that the adjustments to $N$ are accurately reflected in the average number of bytes per chunk. While BPIC is not a direct performance metric, it serves as a vital diagnostic tool for assessing the variance in chunk sizes during the dynamic chunking process. This analysis is crucial for understanding the model’s linguistic abstractions and provides a quantitative measure of how the targeted compression ratio $N$ influences internal semantic segmentation across two distinct training phases:
\begin{itemize}
\item \textbf{Phase 1 (Initial Feature Acquisition):} During early training, we set $N$ to a low range. In this stage, the router is encouraged to be highly sensitive, treating even moderate feature differences as segment boundaries. This results in smaller chunks and a fine-grained representation, minimizing the risk of information loss while the model learns basic byte-level patterns and local context.
\item \textbf{Phase 2 (Compression Transition):} As training progresses, $N$ is gradually increased. This forces the router to become more selective, identifying only significant differences as boundaries. The model begins to learn how to compress information into larger, more efficient blocks, shifting its focus from local byte sequences to broader contextual structures.
\end{itemize}

\section{Ablation Study}
\label{ablation}
In this section, we perform an ablation analysis to evaluate the impact of the compression ratio on hierarchical byte-level models. To manage computational costs, these experiments were conducted using the H-Net 350M variant, trained on a 40\% random subset of the 100B FineWeb-Edu dataset. We investigate whether maintaining a fixed compression ratio $N$ ranging from low ($N=7.0$) and medium ($N=8.0$) to high ($N=9.0$).

The results presented in Table \ref{tab:hnet350_evaluation} and Table \ref{tab:hnet350_hella} confirm that our method of adapting $N$ values from 8.0 up to 9.5 generally outperforms conventional fixed-rate methods across standard benchmarks and robustness tests. While the validation BPB for all models is similar (approximately 0.828) due to the reduced data volume and model size, the downstream tasks provide a clearer demonstration of our method's effectiveness.
\begin{table*}[h]
\centering
\caption{\textbf{Ablation Study.} H-Net 350M trained on 40\% random Fineweb-Edu  variants across different benchmarks}
\label{tab:hnet350_evaluation}
\begin{tabular}{lccccccccc}
\hline
\textbf{MODEL} & \textbf{BPIC} & \textbf{ARC-C} & \textbf{ARC-E} & \textbf{HELLA} & \textbf{LAMB} & \textbf{OPEN} & \textbf{PIQA} & \textbf{WINO} & \textbf{AVERAGE-ACC} \\
\hline
H-Net 350M (base.lower) & 5.11 & 30.1 & 56.4 & \textbf{45.7} & 38.9 & 34.0 & 68.1 & \textbf{54.1} & 46.8 \\
H-Net 350M (base.mid) & 5.42 & 29.4 & 52.7 & 45.0 & 36.6 & 38.0 & \textbf{68.7} & 53.2 & 46.2 \\
H-Net 350M (base.upper) & 5.68 & 31.0 & \textbf{57.0} & 45.6 & \textbf{39.4} & 36.4 & 68.2 & 51.5 & 47.0 \\
H-Net 350M (ours) & 5.51 & \textbf{32.0} & 56.2 & 45.6 & 38.5 & \textbf{37.4} & 68.0 & 53.1 & \textbf{47.3} \\
\hline
\end{tabular}

\label{tab:hnet_results}
\end{table*}

\textbf{Impact of Fixed vs. Adaptive Compression.}

Table \ref{tab:hnet350_evaluation} illustrates the trade-offs inherent in fixed-rate compression. While specific fixed rates show localized strengths, such as base.upper performing well on ARC-Easy (57.0) and LAMBADA (39.4), none of the fixed configurations maintain high performance across all tasks.

In contrast, our method achieves the highest Average Accuracy (47.3\%). Our method demonstrates a superior ability to generalize, particularly in the ARC-Challenge (+1.0\% over the best baseline) and OpenBookQA tasks. This suggests that the adaptive mechanism effectively allocates information density where it is most needed, rather than forcing a uniform sequence length that may discard critical details in complex contexts.

\textbf{Textual perturbations}

We further evaluated the impact of compression on model robustness under textual perturbations, as shown in Table \ref{tab:hnet350_hella}. The performance across the different fixed-rate baselines is relatively stagnant, with average accuracies hovering between 28.8\% and 28.9\%.
Our adaptive method achieves the highest overall robustness score (29.0\%), with a notable peak in the UPPERCASE category (34.5\%). While the margins in the robustness ablation are tighter than in the standard benchmarks, the consistency of our method across Drop and Uppercase perturbations indicates that learning segmentation strategies jointly with the network provides a more stable representation than heuristic-based fixed chunking.

\begin{table*}[h]
\centering
\caption{\textbf{Ablation Study.} Textual Perturbation evaluation for H-Net 350M variants}
\label{tab:hnet350_hella}
\begin{tabular}{lcccccc}
\hline
\textbf{MODEL} & \textbf{ANTSPEAK} & \textbf{DROP} & \textbf{RANDOMCASE} & \textbf{REPEAT} & \textbf{UPPERCASE} & \textbf{AVG-ACC} \\
\hline
H-Net 350M (base.lower) & \textbf{29.2} & 26.9 & \textbf{26.3} & 27.6 & 34.2 & 28.9 \\
H-Net 350M (base.mid) & \textbf{29.2} & 27.2 & \textbf{26.3} & \textbf{27.8} & 34.0 & 28.9 \\
H-Net 350M (base.upper) & 29.1 & 27.1 & 26.2 & 27.6 & 33.8 & 28.8 \\
H-Net 350M (ours) & 29.1 & \textbf{27.3} & \textbf{26.3} & 27.5 & \textbf{34.5} & \textbf{29.0} \\
\hline
\end{tabular}

\label{tab:robustness}
\end{table*}




\section{Conclusion}
\label{conclusion}
In this work, we have addressed the inherent limitations of subword tokenization by proposing ATDC (Adaptive Targeted Dynamic Chunking), a novel byte-level hierarchical framework. Through extensive evaluation across multiple scales from 350M to 1.3B parameters. Our results demonstrate that ATDC consistently outperforms both traditional token-based architectures like Llama 3.2 and state-of-the-art byte-level baselines using fixed compression strategies.

Our empirical analysis yields three primary contributions. First, we demonstrate that byte-level modeling provides superior language modeling quality (BPB) and zero-shot reasoning capabilities compared to token-based models. Notably, our 680M variant outperforms the Llama 3.2 1.5B model, suggesting that direct byte-level processing captures linguistic nuances that subword tokenization often obscures. Second, our robustness evaluations prove that ATDC is significantly more resilient to out-of-distribution noise and textual perturbations. By avoiding the ``tokenization preprocessing", our model maintains high performance in the presence of typos, case variations, and formatting shifts where traditional models operate worse (Table \ref{tab:hellag_perturbation}).

Finally, our ablation studies and boundary visualizations confirm that adaptive targeted chunking is essential for effective hierarchical modeling. Unlike fixed-rate compression, which often fragments semantic units, our adaptive mechanism learns to align chunk boundaries with meaningful morphological structures. This allow the model to dynamically allocate information density, leading to more stable representations and improved generalization across complex reasoning tasks. 

\section*{Acknowledgment}
LLMs were used solely to assist with proofreading. All research ideas, theoretical development, experiments, and implementation were carried out entirely by the authors.
\bibliographystyle{IEEEtran}
\bibliography{mybibliography}

@inproceedings{hwang2025dynamic,
title={Dynamic Chunking for End-to-End Hierarchical Sequence Modeling},
author={Hwang, Sukjun and Wang, Brandon and Gu, Albert},
booktitle={The 14th ICLR},
year={2026}
}

@inproceedings{nawrot2022hierarchical,
  title={Hierarchical transformers are more efficient language models},
  author={Nawrot, Piotr and Tworkowski, Szymon and Tyrolski, Micha{\l} and Kaiser, {\L}ukasz and Wu, Yuhuai and Szegedy, Christian and Michalewski, Henryk},
  booktitle={Findings of the ACL: NAACL 2022},
  pages={1559--1571},
  year={2022}
}

@article{grattafiori2024llama,
  title={The llama 3 herd of models},
  author={Grattafiori, Aaron and Dubey, Abhimanyu and Jauhri, Abhinav and Pandey, Abhinav and Kadian, Abhishek and Al-Dahle, Ahmad and Letman, Aiesha and Mathur, Akhil and Schelten, Alan and Vaughan, Alex and others},
  journal={arXiv preprint arXiv:2407.21783},
  year={2024}
}

@inproceedings{neitemeier2025hierarchical,
title={Hierarchical Autoregressive Transformers: Combining Byte- and Word-Level Processing for Robust, Adaptable Language Models},
author={Pit Neitemeier and Bj{\"o}rn Deiseroth and Constantin Eichenberg and Lukas Balles},
booktitle={The 13th ICLR},
year={2025}
}

@inproceedings{pagnoni2025byte,
  title={Byte latent transformer: Patches scale better than tokens},
  author={Pagnoni, Artidoro and Pasunuru, Ramakanth and Rodriguez, Pedro and Nguyen, John and Muller, Benjamin and Li, Margaret and Zhou, Chunting and Yu, Lili and Weston, Jason E and Zettlemoyer, Luke and others},
  booktitle={Proceedings of the 63rd the ALC (Volume 1: Long Papers)},
  pages={9238--9258},
  year={2025}
}

@article{bengio2013estimating,
  title={Estimating or propagating gradients through stochastic neurons for conditional computation},
  author={Bengio, Yoshua and L{\'e}onard, Nicholas and Courville, Aaron},
  journal={arXiv preprint arXiv:1308.3432},
  year={2013}
}

@InProceedings{pmlr-v235-dao24a,
  title = 	 {Transformers are {SSM}s: Generalized Models and Efficient Algorithms Through Structured State Space Duality},
  author =       {Dao, Tri and Gu, Albert},
  booktitle = 	 {Proceedings of the 41st ICML},
  pages = 	 {10041--10071},
  year = 	 {2024},
  volume = 	 {235}
}

@article{penedo2024fineweb,
  title={The fineweb datasets: Decanting the web for the finest text data at scale},
  author={Penedo, Guilherme and Kydl{\'\i}{\v{c}}ek, Hynek and Lozhkov, Anton and Mitchell, Margaret and Raffel, Colin A and Von Werra, Leandro and Wolf, Thomas and others},
  journal={Advances in NeurIPS},
  volume={37},
  pages={30811--30849},
  year={2024}
}

@article{loshchilov2017decoupled,
  title={Decoupled weight decay regularization},
  author={Loshchilov, Ilya and Hutter, Frank},
  journal={arXiv preprint arXiv:1711.05101},
  year={2017}
}

@article{hu2024minicpm,
  title={Minicpm: Unveiling the potential of small language models with scalable training strategies},
  author={Hu, Shengding and Tu, Yuge and Han, Xu and He, Chaoqun and Cui, Ganqu and Long, Xiang and Zheng, Zhi and Fang, Yewei and Huang, Yuxiang and Zhao, Weilin and others},
  journal={arXiv preprint arXiv:2404.06395},
  year={2024}
}

@article{ibrahim2024simple,
title={Simple and Scalable Strategies to Continually Pre-train Large Language Models},
author={Adam Ibrahim and Benjamin Th{\'e}rien and Kshitij Gupta and Mats Leon Richter and Quentin Gregory Anthony and Eugene Belilovsky and Timoth{\'e}e Lesort and Irina Rish},
journal={TMLR},
issn={2835-8856},
year={2024}
}

@article{xue2022byt5tokenfreefuturepretrained,
  title={ByT5: Towards a token-free future with pre-trained byte-to-byte models},
  author={Xue, Linting and Barua, Aditya and Constant, Noah and Al-Rfou, Rami and Narang, Sharan and Kale, Mihir and Roberts, Adam and Raffel, Colin},
  journal={Transactions of the ACL},
  volume={10},
  pages={291--306},
  year={2022},
}

@inproceedings{gu2022efficiently,
title={Efficiently Modeling Long Sequences with Structured State Spaces},
author={Albert Gu and Karan Goel and Christopher Re},
booktitle={International Conference on Learning Representations},
year={2022}
}

@inproceedings{yu2023megabyte,
title={{MEGABYTE}: Predicting Million-byte Sequences with Multiscale Transformers},
author={LILI YU and Daniel Simig and Colin Flaherty and Armen Aghajanyan and Luke Zettlemoyer and Mike Lewis},
booktitle={Thirty-seventh Conference on NeurIPS},
year={2023}
}

@inproceedings{wang2024mambabyte,
title={MambaByte: Token-free Selective State Space Model},
author={Junxiong Wang and Tushaar Gangavarapu and Jing Nathan Yan and Alexander M Rush},
booktitle={First Conference on Language Modeling},
year={2024}
}

@inproceedings{kudo2018sentencepiece,
  title={SentencePiece: A simple and language independent subword tokenizer and detokenizer for Neural Text Processing},
  author={Kudo, Taku and Richardson, John},
  booktitle={Proceedings of the 2018 Conference on EMNLP},
  pages={66--71},
  year={2018}
}

@inproceedings{song2021fast,
  title={Fast wordpiece tokenization},
  author={Song, Xinying and Salcianu, Alex and Song, Yang and Dopson, Dave and Zhou, Denny},
  booktitle={Proceedings of the 2021 conference on empirical methods in natural language processing},
  pages={2089--2103},
  year={2021}
}

@article{petrov2023language,
  title={Language model tokenizers introduce unfairness between languages},
  author={Petrov, Aleksandar and La Malfa, Emanuele and Torr, Philip and Bibi, Adel},
  journal={Advances in NeurIPS},
  volume={36},
  pages={36963--36990},
  year={2023}
}

@inproceedings{ahia2023do,
title={Do All Languages Cost the Same? Tokenization in the Era of Commercial Language Models},
author={Orevaoghene Ahia and Sachin Kumar and Hila Gonen and Jungo Kasai and David R Mortensen and Noah A. Smith and Yulia Tsvetkov},
booktitle={The 2023 Conference on Empirical Methods in Natural Language Processing},
year={2023}
}

@inproceedings{schmidt2025boundless,
title={Boundless Byte Pair Encoding: Breaking the Pre-tokenization Barrier},
author={Craig W Schmidt and Varshini Reddy and Chris Tanner and Yuval Pinter},
booktitle={Second CLM},
year={2025}
}

@article{shazeer2020glu,
  title={Glu variants improve transformer},
  author={Shazeer, Noam},
  journal={arXiv preprint arXiv:2002.05202},
  year={2020}
}

@article{brotons2024exponential,
title={Exponential Moving Average of Weights in Deep Learning: Dynamics and Benefits},
author={Daniel Morales-Brotons and Thijs Vogels and Hadrien Hendrikx},
journal={Transactions on Machine Learning Research},
issn={2835-8856},
year={2024}
}

@article{zakershahrak2025h,
  title={H-Net++: Hierarchical Dynamic Chunking for Tokenizer-Free Language Modelling in Morphologically-Rich Languages},
  author={Zakershahrak, Mehrdad and Ghodratnama, Samira},
  journal={arXiv preprint arXiv:2508.05628},
  year={2025}
}

@inproceedings{dang2022regularizing,
  title={Regularizing Data for Improving Execution Time of NLP Model},
  author={Dang, Thang and Sakai, Yasufumi and Tabaru, Tsuguchika and Kasagi, Akihiko},
  booktitle={The International FLAIRS Conference Proceedings},
  volume={35},
  year={2022}
}

@inproceedings{kasagi2021efficient,
  title={Efficient and large scale pre-training techniques for Japanese natural language processing},
  author={Kasagi, Akihiko and Asaoka, Masahiro and Tabuchi, Akihiro and Oyama, Yosuke and Honda, Takumi and Sakai, Yasufumi and Dang, Thang and Tabaru, Tsuguchika},
  booktitle={2021 Ninth International Symposium on Computing and Networking (CANDAR)},
  pages={108--113},
  year={2021},
  organization={IEEE}
}
\end{document}